\documentclass[pdflatex,sn-mathphys-num]{sn-jnl}

\usepackage{graphicx}%
\usepackage{multirow}%
\usepackage{amsmath,amssymb,amsfonts}%
\usepackage{amsthm}%
\usepackage{mathrsfs}%
\usepackage[title]{appendix}%
\usepackage{xcolor}%
\usepackage{textcomp}%
\usepackage{manyfoot}%
\usepackage{booktabs}%
\usepackage{algorithm}%
\usepackage{algorithmicx}%
\usepackage{algpseudocode}%
\usepackage{listings}%
\usepackage{url}
\usepackage{graphicx}
\usepackage{subcaption}

\theoremstyle{thmstyleone}%
\theoremstyle{thmstyletwo}%

\theoremstyle{thmstylethree}%

\begin{document}

\title[Article Title]{Hyperparameter and Neural Architecture Search via Evolutionary Optimization for Reinforcement Learning: HPNAS-RL}


\author*[1,2]{\fnm{Nihal} \sur{Acharya Adde$^{[0009-0002-0966-0825]}$}}\email{nihal.acharya.adde@campus.tu-berlin.de, nihal.acharya.adde@volkswagen.de}

\author[2]{\fnm{Hanno} \sur{Gottschalk$^{[0000-0003-2167-2028]}$}}\email{gottschalk@math.tu-berlin.de}

\affil*[1]{\orgdiv{Volkswagen Group Innovation}, \orgname{Volkswagen AG}, \orgaddress{\street{Berliner Ring 2}, \city{Wolfsburg}, \postcode{38440}, \state{Lower Saxony}, \country{Germany}}}

\affil[2]{\orgdiv{Institute of Mathematics}, \orgname{TU Berlin}, \orgaddress{\street{Straße des 17. Juni 136}, \city{Berlin}, \postcode{10623}, \state{Berlin}, \country{Germany}}}


\abstract{In this study, we propose a dual-branch evolutionary framework that simultaneously performs Neural Architecture Search (NAS) and Hyperparameter Optimization (HPO) for Reinforcement Learning (RL) in Autonomous Driving (AD) tasks. While NAS has shown strong potential in automating architecture design for Convolutional Neural Networks (CNNs), its application to RL remains limited, especially in scenarios that require joint optimization of architectural and training dynamics. Building on our prior work on Evolutionary Multi-Objective NAS (EMNAS), the proposed framework evolves discrete network architectures and continuous hyperparameters in parallel under a unified multi-objective search strategy. In this setup, NAS targets the convolutional layers of the RL policy network - responsible for visual feature extraction, while HPO focuses on optimizing critical RL hyperparameters. To accelerate convergence and improve generalization, we incorporate parallel evaluation and a teacher-student policy transfer mechanism. Experiments in a high-fidelity driving simulator show that the joint NAS-HPO framework consistently outperforms standalone NAS, yielding policies with higher rewards, fewer parameters, and improved stability. These results highlight the benefits of jointly optimizing model structure and training dynamics for scalable and effective RL agent design.}




\maketitle

\section{Introduction}\label{sec1}
Reinforcement Learning (RL) has emerged as a powerful framework for Autonomous Driving (AD), enabling agents to learn sophisticated decision-making and control policies in complex environments. However, designing effective RL agents requires carefully crafting both the neural network architecture and the associated training hyperparameters, which together determine the agent's performance, sample efficiency, and computational cost. Traditionally, these design decisions are made manually, relying on expert intuition and extensive trial-and-error tuning, which are time-consuming and often suboptimal.

To automate and scale this process, we previously introduced the Evolutionary Multi-Objective Neural Architecture Search (EMNAS) framework \cite{adde2025emnas}, building on EEEANet \cite{termritthikun2021eeea}. EMNAS employs a Genetic Algorithm (GA) to evolve compact convolutional architectures for the policy network of a Proximal Policy Optimization (PPO) based RL agent. By integrating early-exit evaluation, Optimized Transfer Learning (OTL), and parallelized training, EMNAS demonstrated superior performance over manually crafted models \cite{adde2025emnas}. However, its search space was limited solely to architectural configurations, neglecting the significant impact of RL hyperparameters on RL convergence and stability.

In this paper, we extend EMNAS into a dual-branch evolutionary framework that jointly optimizes neural network architectures and training hyperparameters. Hyperparameter and Neural Architecture Search (HPNAS) framework combines discrete architecture evolution with continuous hyperparameter tuning using polynomial mutation and Simulated Binary Crossover (SBX), enabling coordinated search across both spaces. Neural Architecture Search (NAS) targets the convolutional layers of the RL policy network, which are responsible for visual feature extraction, while Hyperparameter Optimization (HPO) tunes RL hyperparameters such as the learning rate, batch size, discount factor and time horizon. Initial experiments demonstrate that the HPNAS strategy yields consistent improvements in cumulative reward, model compactness, and training stability over architecture-only baselines. A Bayesian surrogate optimization model is being developed to learn the relationship between architecture and hyperparameter configurations and their performance. Although not yet integrated in this work, it is designed to guide later stage search by predicting promising configurations and is discussed further in the outlook.

\subsection{Autonomous Driving Task:}
\label{sec:ad_setup}
The AD task \cite{10.1007/978-3-031-82484-5_3} is performed in a simulated environment built using the Unity 3D engine \cite{Unity}, where multiple agents are trained with the Proximal Policy Optimization (PPO) algorithm \cite{schulman2017proximal}. Agents navigate a three-lane oval track with the goal of learning smooth, collision-free driving while avoiding abrupt maneuvers. Each agent receives semantic segmentation images from a forward-facing camera and predicts future trajectory points, which are executed by an external black-box vehicle controller responsible for throttle, brake, and steering commands. The reward function promotes accurate trajectory prediction, penalizes collisions and lane deviations, and encourages stable motion with minimal jerk. 

\begin{figure}[H]
    \centering
    \includegraphics[width=0.9\linewidth]{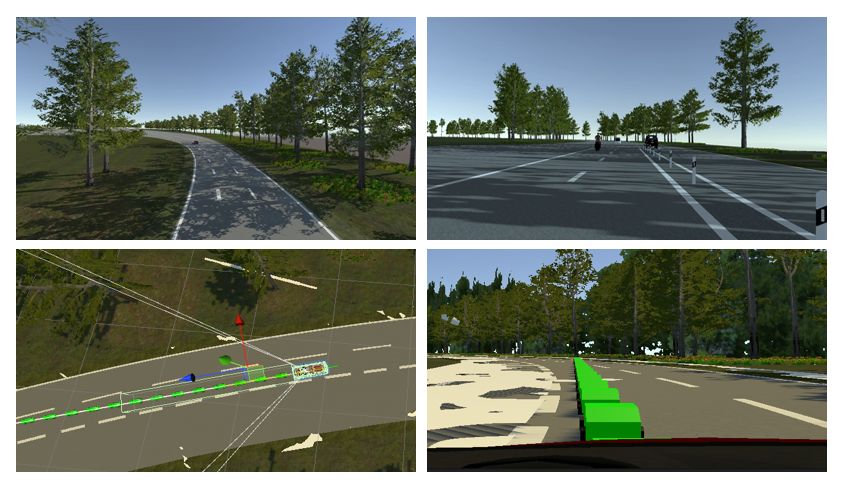}
    \caption{Simulated three-lane track used for RL-based AD training in Unity3D. Top left: Drone view of road with agent. Top right: Front camera view, later converted to a semantic segmentation image and used as the network input. Bottom: Driving behavior based on predicted trajectory points.}
    \label{fig:unity_env}
\end{figure}

Figure~\ref{fig:unity_env} illustrates the simulated Unity3D environment used for agent training. This setup provides a realistic yet controlled platform for evaluating reinforcement learning strategies and their generalization across training runs. In this work, we focus on optimizing the convolutional network architecture and PPO hyperparameters underlying the policy to enhance driving performance and training efficiency in the RL-based AD task.

\section{Related Work}

NAS has evolved through RL-based search~\cite{zoph2016neural}, gradient-based optimization~\cite{liu2018darts}, and Evolutionary Algorithms (EAs)~\cite{real2019regularized}. While gradient-based methods such as DARTS~\cite{liu2018darts} improve efficiency by making architecture selection differentiable through continuous weighting of candidate operations, they often converge to locally optimal designs and are restricted to differentiable objectives. RL-based NAS frameworks~\cite{zoph2016neural} have shown promising results but require extensive computational resources due to sequential policy evaluations. In contrast, EAs provide a more flexible search paradigm capable of exploring large, non-differentiable spaces while maintaining population diversity, making them well-suited for complex architecture optimization problems. Early methods such as NeuroEvolution of Augmenting Topologies (NEAT)~\cite{stanley2002NEAT} demonstrated the ability of genetic algorithms to evolve both network topology and parameters. Later studies, including AmoebaNet~\cite{real2019regularized}, showed that evolutionary search can outperform reinforcement learning and random search strategies in large-scale vision tasks. Building on these ideas, Multi-Objective Evolutionary Algorithms (MOEAs) were introduced to balance accuracy, computational cost, and network size. Examples include NSGA-Net~\cite{lu2019nsga}, which generates a Pareto front of efficient architectures, and Early Exit Evolutionary Neural Architecture Search (EEEANet)~\cite{termritthikun2021eeea}, which combines MOEAs with early-exit initialization using the NSGA-II algorithm~\cite{deb2002fast} to improve convergence. Despite their effectiveness, most EA and MOEA-based NAS frameworks remain confined to supervised learning, offering limited scalability and adaptability to reinforcement learning tasks where sample efficiency and stability are critical.

In recent years, researchers have recognized that architectural choices and hyperparameter settings are interdependent, motivating the development of joint optimization approaches that integrate NAS with HPO~\cite{hirose2021nashpobench2, bansal2022jahs}. Such methods have shown that jointly tuning both components can lead to more stable and higher-performing models, as hyperparameters directly influence architecture behavior during training. Bayesian optimization-based strategies~\cite{kandasamy2018neural} and weight-sharing techniques~\cite{li2021random} have been proposed to accelerate search by reusing knowledge across configurations. Evolutionary approaches for joint NAS–HPO have also been explored in convolutional network settings~\cite{bria2023evo,guerrero2021bag}. These methods demonstrated the potential of integrating architectural and hyperparameter optimization, using evolutionary multi-objective strategies such as non-dominated sorting and tournament selection to balance competing objectives. However, they were primarily designed for static supervised learning tasks, where evaluations are deterministic and data distributions remain fixed. Such frameworks do not address the unique challenges of reinforcement learning, including stochastic rewards, non-stationary environments, and the need for scalable exploration across parallel agents. 

In this context, our work introduces a dual-branch evolutionary framework that jointly optimizes architectures and RL-specific hyperparameters. By combining a Directed Acyclic Graph (DAG)-based NASNet-style representation~\cite{zoph2018learning} with parallel population training and transfer learning mechanisms, the proposed approach achieves scalable and sample-efficient optimization in high-dimensional RL environments, bridging the gap between HPO, NAS and RL.

\section{HPNAS Methodology}
The proposed HPNAS framework extends our earlier EMNAS system \cite{adde2025emnas} by introducing a dual-branch evolutionary approach that jointly optimizes neural architectures and reinforcement learning hyperparameters. As illustrated in Sections~\ref{sec:emnas_framework} and~\ref{sec:nas_hpo}, the framework comprises two coordinated components: the EMNAS-based architecture optimization branch, which evolves convolutional network structures, and the hyperparameter optimization branch, which tunes PPO-related parameters such as learning rate, batch size, and discount factor. Both branches operate within a unified multi-objective evolutionary process, sharing fitness evaluations to jointly enhance policy performance and model efficiency.

\subsection{EMNAS Framework}
\label{sec:emnas_framework}

\begin{figure}[!ht]
\centering
    \includegraphics[width=\linewidth, height=100pt]{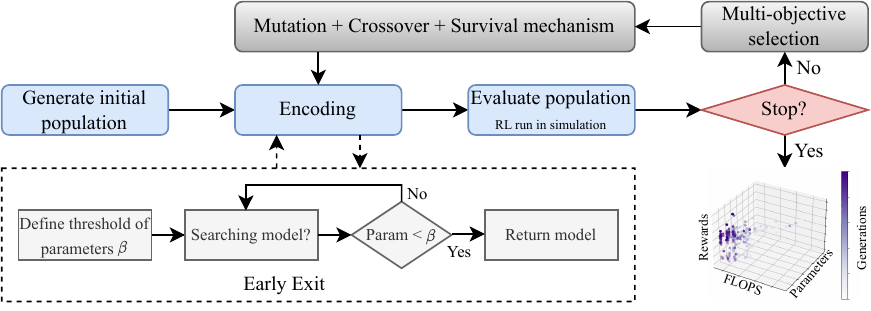}
 \caption{Overview of the EMNAS pipeline. Multi-objective evolutionary optimization combines Early-Exit Population Initialization (EEPI) and Optimized Transfer Learning (OTL) for efficient RL architecture discovery.}
\label{fig:emnas_framework}
\end{figure}

Our previous work, EMNAS \cite{adde2025emnas}, builds upon the EEEANet framework \cite{termritthikun2021eeea} to optimize convolutional neural architectures for RL tasks based on the PPO algorithm. Using a multi-objective genetic algorithm, EMNAS evolves candidate architectures to simultaneously maximize PPO cumulative reward while minimizing computational cost and model complexity. The optimization problem is formulated as:
\begin{equation} \label{eq:1}
    f(x) = \min \left\{ -\textbf{Reward}(x), \textbf{FLOPS}(x), \textbf{Params}(x) \right\}
\end{equation}
where $\textbf{Reward}(x)$ denotes the PPO training reward, $\textbf{FLOPS}(x)$ represents the floating-point operations per second, and $\textbf{Params}(x)$ indicates the number of learnable parameters in the network. All objectives are minimized to identify models that achieve a favorable balance between accuracy and efficiency. An overview of the EMNAS pipeline, including the genotype encoding and evolutionary process, is shown in Figure~\ref{fig:emnas_framework}.

\subsubsection{Genotype Encoding Scheme}
Architectures in EMNAS are represented as chromosomes composed of alternating normal and reduction cells \cite{adde2025emnas,zoph2018learning}. A normal cell preserves the spatial resolution of the input feature map, while a reduction cell halves it to enable hierarchical feature extraction. Each cell consists of several genes, defined as $LA_{1}LA_{2}$, $LB_{1}LB_{2}$, $LC_{1}LC_{2}$, and $LD_{1}LD_{2}$, where each gene specifies a convolutional operation $L$ and its corresponding connection indices $(A,B,C,D)$:
\begin{equation}
\text{chromosome}(x) = LA_{1}LA_{2},LB_{1}LB_{2},LC_{1}LC_{2},LD_{1}LD_{2}
\label{eq:chromosome_encoding}
\end{equation}
Here, $L$ represents the operator selected from the predefined search space, while indices $(A,B,C,D)$ determine the connectivity pattern between feature maps. The operation set includes depth-wise separable, dilated, and inverted convolutions, pooling operations, and skip connections as shown in Table~\ref{tab:search_space}.

\begin{table}[h]
\centering
\caption{Search space used for normal and reduction cells in EMNAS.}
\label{tab:search_space}
\begin{tabular}{@{}ll@{}}
\toprule
\textbf{Operation Type} & \textbf{Details} \\ \midrule
\textit{Convolutional Operations} &
3×3 depth-wise separable convolution \\
& 5×5 depth-wise separable convolution \\
& 3×3 dilated convolution \\
& 5×5 dilated convolution \\
& 3×3 inverted convolution \\
& 5×5 inverted convolution \\
& 7×7 convolution \\[3pt]
\textit{Pooling Operations} &
Max pooling \\
& Average pooling \\[3pt]
\textit{Other Operations} &
Skip connection \\ \bottomrule
\end{tabular}
\end{table}

The connections of each cell are defined according to the index positions $A, B, C, D$, which indicate where the outputs of previous operations are concatenated. For instance, assuming nine operations indexed as $[0,1,2,3,4,5,6,7,8]$, a gene $LA_{1}LA_{2}=30,80$ implies that outputs from operations 3 and 8 are connected at index 0, while $LB_{1}LB_{2}=01,21$ connects operations 0 and 2 at index 1. Genes $LC_{1}LC_{2}$ and $LD_{1}LD_{2}$ connect sequentially to outputs of earlier nodes to maintain acyclic connectivity and reduce redundant parameters.

The index of each connection is computed recursively as:
\begin{equation}
\text{PreviousIndex} = \text{index} - 2
\label{eq:previous_index}
\end{equation}
The \textit{PreviousIndex} defines the node from which the current operation receives its input, ensuring that connections always originate from earlier nodes. For example, a gene $LC_{1}LC_{2}=12,02$ (index = 2) connects from index 0, while $LD_{1}LD_{2}=63,73$ (index = 3) connects from index 1. If different indices exist within a gene, such as $LD_{1}LD_{2}=63,72$, operation 6 connects from index 1 and operation 7 from index 0. This hierarchical indexing maintains directed acyclic structure and enforces valid feature flow within each cell.

This compact encoding jointly captures both the architectural structure and inter-layer dependencies, enabling EMNAS to perform efficient, flexible search over diverse CNN cell configurations.

\subsubsection{Multi-objective Evolutionary Algorithm}
The optimization of CNN architectures in EMNAS is formulated as a multi-objective problem, aiming to maximize reinforcement learning performance while minimizing computational cost and model complexity:
\begin{equation}\label{eq:moop_general}
\begin{aligned}
\min & \quad f_1(x), f_2(x), \dots, f_k(x) \\
\text{s.t.} & \quad x \in X
\end{aligned}
\end{equation}
In EMNAS, $k=3$, and the objective vector is defined as in Equation~\ref{eq:1}. A population of candidate architectures is evolved over multiple generations, where each individual is trained and evaluated using the defined objectives. Fitness values are then ranked using the NSGA-II algorithm \cite{deb2002fast}, which applies non-dominated sorting and crowding distance to preserve diversity among Pareto-optimal solutions. This enables the search to discover a balanced set of architectures that achieve high PPO reward while remaining lightweight and computationally efficient. Equal weighting is applied to all objectives to ensure fair comparison across the trade-off surface.

\subsubsection{Genetic Operations}
Each architecture in EMNAS is represented as a Directed Acyclic Graph (DAG) defined by $G = (V, E)$, where $V$ denotes the set of nodes corresponding to convolutional operators and $E$ represents the directed edges encoding their input connections. Offspring are generated through crossover, mutation, and survival operations, which introduce diversity while preserving the structural validity of the architecture.

\paragraph{Crossover:}
Uniform crossover combines two parent architectures
\(G_1=(V_1,E_1)\) and \(G_2=(V_2,E_2)\) to form a child
\(G_c=(V_c,E_c)\). Each gene \((v_i,e_i)\) is inherited from either
parent with probability \(p_c=0.5\):
\begin{equation}
  (v_i^{c}, e_i^{c}) =
  \begin{cases}
    (v_i^{1}, e_i^{1}), & \text{if } r_i < p_c,\\
    (v_i^{2}, e_i^{2}), & \text{otherwise,}
  \end{cases}
\end{equation}
where $r_i \sim U(0,1)$. This operation promotes genetic mixing by combining convolutional operations and their input connections from both parents, generating offspring that inherit key structural traits while introducing novel layer combinations.

\paragraph{Mutation:}
With probability \(p_m\), a randomly selected gene \((v_i, e_i)\) is replaced by
\((v_i', e_i')\) drawn from the predefined index and operation sets:
\begin{equation}
  (v_i', e_i') =
  \begin{cases}
    \text{Random}(V,E), & \text{with probability } p_m,\\
    (v_i, e_i), & \text{otherwise.}
  \end{cases}
\end{equation}
Additionally, an inversion mutation with probability $p_{inv}$ reverses the order of a randomly selected subgraph while maintaining the DAG structure. Typical values are $p_m = 0.1$ and $p_{inv} = 0.1$. Mutation introduces controlled variability in the population, enhancing exploration while preventing premature convergence.

\paragraph{Survival:}
A survival mechanism ensures that the best-performing architectures from the current generation are retained in the next. This strategy preserves strong solutions to maintain knowledge continuity while allowing new offspring to introduce diversity. By combining exploitation of high-fitness models with exploration of novel candidates, the population converges more effectively toward the Pareto-optimal frontier.

\paragraph{Early Exit Parameter Initialization (EEPI):}
EMNAS applies a maximum parameter constraint $\beta_{\max}$ during initialization to avoid oversized models:
\begin{equation}
\text{EEPI}(\theta) =
\begin{cases} 
1, & \text{if } \beta \leq \beta_{\max} \\
0, & \text{otherwise}
\end{cases}
\end{equation}
Architectures exceeding this limit are discarded and reinitialized, which improves early-stage efficiency by reducing computational overhead and restricting the search space to feasible model configurations.

\subsubsection{Optimized Transfer Learning (OTL)}
To improve sample efficiency and ensure stable convergence across generations, EMNAS employs an Optimized Transfer Learning (OTL) mechanism. In this approach, student architectures inherit weights and policies from the best-performing model of the previous generation through behavior cloning \cite{inproceedings}, allowing new populations to begin training from a well-initialized policy rather than from scratch.

At the end of each generation, expert trajectories
$\mathcal{D}_E=\{(s_t,a_t)\}_{t=1}^{T}$ are collected from the top policy
$\pi_E(a\mid s)$, where $s_t\in\mathcal{S}$ is the state (or observation) at time $t$,
$a_t\in\mathcal{A}$ is the action, and $T$ is the trajectory length. Each architecture in
the next generation is initialized by supervised imitation of this expert using the
behavior cloning loss:
\begin{equation}
\mathcal{L}_{BC}(\theta) = \mathbb{E}_{(s,a) \sim \mathcal{D}_E} \left[-\log \pi_\theta(a|s)\right]
\end{equation}
This ensures that both new offspring and surviving individuals begin PPO training from a consistent policy baseline, maintaining a fair optimization trajectory across the population. OTL also retains the optimizer state from the previous generation, including the final learning rate and PPO clipping value, ensuring a smooth continuation of training. By reusing expert data and policy initialization, OTL reduces PPO training time while maintaining stability and fairness across evolving architectures.

\medskip
In summary, EMNAS combines a compact genotype encoding, NSGA-II based multi-objective optimization, and efficiency mechanisms (EEPI and OTL) to discover CNN architectures that jointly optimize RL performance and computational cost. This forms the basis for the dual-branch HPNAS extension.

\subsection{Dual-Branch Extension: Joint NAS and HPO}
\label{sec:nas_hpo}

While EMNAS originally targeted architecture search, RL performance is also strongly influenced by hyperparameter choices. We therefore extend EMNAS into a dual-branch framework that jointly evolves architectures and hyperparameters using a coupled genetic algorithm.

\subsubsection{Hyperparameter Search Space}
Based on the hyperparameter optimization and sensitivity analysis presented in Chapter~\ref{ch:hpo}\cite{10.1007/978-3-031-82484-5_3}, we initially considered six RL hyperparameters for optimization. To reduce the dimensionality and computational cost of the joint search, PPO epochs and the entropy coefficient were fixed at 3 and 0.01, respectively. The remaining four: batch size, discount factor, learning rate, and time horizon form the HPO search space in this work, as they contributed most to performance variation \cite{10.1007/978-3-031-82484-5_3}.The final set of selected hyperparameters and their corresponding ranges are summarized in Table~\ref{tab:hpo_params}. The chosen ranges are informed by previous studies in PPO-based reinforcement learning~\cite{schulman2017proximal,henderson2018deep} and empirical observations in AD environments, ensuring coverage of commonly effective values while maintaining search feasibility.

\begin{table}[h]
\centering
\caption{Tuned Hyperparameters and Ranges}
\label{tab:hpo_params}
\begin{tabular}{|l|c|c|}
\hline
\textbf{Hyperparameter} & \textbf{Range} & \textbf{Description} \\
\hline
Batch Size & $64 - 512$ & Number of samples per gradient update \\
Discount Factor ($\gamma$) & $0.95 - 0.999$ & Weighting of future rewards \\
Learning Rate & $10^{-5} - 10^{-2}$ & Step size for gradient updates \\
Time Horizon & $32 - 256$ & Number of steps per trajectory segment \\
\hline
\end{tabular}
\end{table}

\noindent Our dual-branch framework jointly evolves network architectures and hyperparameters, employing a discrete encoding for NAS and a continuous representation for HPO. Both branches are evaluated using the same multi-objective fitness function as described in EMNAS (Section~\ref{sec:emnas_framework}), which balances reward, FLOPs, and parameter count. In the initial generation, hyperparameters are uniformly sampled within predefined ranges (see Table~\ref{tab:hpo_params}). From the second generation onward, the best-performing configuration is retained through a survival mechanism, and from the remaining population, offspring are generated using crossover or mutation with equal probability (0.5 each). To maintain diversity, 10\% of new individuals are sampled purely at random.

\subsubsection{Crossover and Mutation Mechanisms}
\label{subsec:crossover_mutation}

The continuous hyperparameter space is evolved using Simulated Binary Crossover (SBX) and polynomial mutation, well established operators in multi-objective evolutionary optimization due to their effectiveness in balancing exploration and exploitation~\cite{deb1995simulated,deb2002fast}.

\paragraph{Polynomial mutation:}
Polynomial mutation introduces small stochastic perturbations to continuous parameters, maintaining a balance between local refinement and global exploration. For each hyperparameter \( h_i \), the mutated value \( h_i' \) is computed as:
\begin{equation}
h_i' = h_i + \delta_i (h_i^{\text{max}} - h_i^{\text{min}}),
\end{equation}
where \( \delta_i \) is a mutation offset sampled from a polynomial probability distribution~\cite{deb1995simulated}:
\begin{equation}
\delta_i =
\begin{cases}
(2u)^{\frac{1}{\eta_m + 1}} - 1, & \text{if } u < 0.5,\\[6pt]
1 - [2(1 - u)]^{\frac{1}{\eta_m + 1}}, & \text{otherwise,}
\end{cases}
\end{equation}
with \( u \sim \mathcal{U}(0,1) \) being a uniformly distributed random number, and \( \eta_m \) representing the mutation distribution index.

A larger \( \eta_m \) produces smaller perturbations centered around the parent value \( h_i \), promoting fine-grained local search, whereas a smaller \( \eta_m \) increases the likelihood of larger deviations, enhancing exploration. The scaling term \( (h_i^{\text{max}} - h_i^{\text{min}}) \) ensures that the mutation magnitude is proportional to the allowable range of the hyperparameter. After mutation, values are typically clipped to the interval \([h_i^{\text{min}}, h_i^{\text{max}}]\) to maintain feasibility. This formulation makes polynomial mutation particularly effective for fine-tuning continuous design variables in evolutionary algorithms.

\paragraph{Simulated Binary Crossover (SBX):}
Simulated Binary Crossover (SBX) is a widely used recombination operator for continuous variables that mimics the behavior of single-point crossover in binary-coded genetic algorithms. For each hyperparameter \( h_i \), SBX generates two offspring \( h_i' \) and \( h_i'' \) from two parent values \( h_i^1 \) and \( h_i^2 \) as:
\begin{equation}
\begin{aligned}
h_i'  &= 0.5 \left[ (1 + \beta_i) h_i^1 + (1 - \beta_i) h_i^2 \right],\\[4pt]
h_i'' &= 0.5 \left[ (1 - \beta_i) h_i^1 + (1 + \beta_i) h_i^2 \right],
\end{aligned}
\end{equation}
where \( \beta_i \) is a spread factor sampled according to the SBX probability distribution~\cite{deb1995simulated}:
\begin{equation}
\beta_i =
\begin{cases}
(2u)^{\frac{1}{\eta_c + 1}}, & \text{if } u \leq 0.5,\\[6pt]
\left[\frac{1}{2(1 - u)}\right]^{\frac{1}{\eta_c + 1}}, & \text{otherwise,}
\end{cases}
\end{equation}
with \( u \sim \mathcal{U}(0,1) \) being a uniformly distributed random number, and \( \eta_c \) denoting the crossover distribution index.

A larger \( \eta_c \) produces offspring closer to the parents, promoting exploitation, while a smaller \( \eta_c \) allows for a wider spread of solutions, enhancing exploration. The symmetric nature of SBX ensures that both offspring inherit information from both parents, maintaining diversity while preserving continuity in the search space. This property makes SBX particularly effective for real-coded evolutionary algorithms applied to hyperparameter optimization, where smooth variation between parent configurations enables efficient exploration of the search space.

\medskip
For the architecture branch, we retain EMNAS’s discrete mutation and crossover \cite{adde2025emnas}. Together, these operators enable efficient joint search over architecture and training dynamics, supporting scalable and effective RL agent discovery.

\section{Experimental Setup}
We evaluate the proposed dual-branch evolutionary framework across multiple autonomous driving experiments, designed to test its effectiveness in jointly optimizing architectures and hyperparameters under resource constrained RL settings. The experiments are conducted in the Unity3D simulation environment described in Section~\ref{sec:ad_setup}, where multiple agents are trained using the PPO algorithm. Each candidate is evaluated with a multi-objective fitness that balances policy reward, FLOPs, and parameter count, consistent with EMNAS~\cite{adde2025emnas}.

The evolutionary algorithm employs a crossover probability of 0.5 and a mutation probability of 0.1 for the architecture branch, while the hyperparameter branch uses a higher mutation probability of 0.5 to encourage exploration in continuous search spaces. A survival rate of 20\% (up to four top-performing individuals retained per generation) is applied across both branches. To maintain diversity, 10\% of offspring are generated through uniform random sampling. We schedule both the crossover and mutation probabilities, and the per-generation distribution index of crossover and mutation, to shift from exploration in early generations to exploitation in later ones. Candidate policies are trained and evaluated under low-fidelity settings to reduce computational cost, including 40 PPO epochs, input resolution of $84 \times 84 \times 3$, and a network depth of 8. The depth corresponds to eight stacked cells alternating between normal and reduction cells, where the discovered architectures define the internal operations and connections within each cell type. These cells are then repeated to construct the final policy network. As we employ relative ranking within each generation, evaluation bias is reduced despite the low-fidelity settings. The architecture search is bounded using an EEPI threshold of 5 million parameters~\cite{adde2025emnas}. As in prior work, we employ an optimized transfer learning strategy where student models are initialized via behavior cloning \cite{inproceedings} from the best policy of the previous generation~\cite{adde2025emnas}.

We run six experiments with the following (generations, population size) settings: (6, 6), (10, 4), (10, 10), (10, 15), (15, 15), (15, 20). All evaluations are parallelized across NVIDIA Tesla H100 GPUs. We benchmark against the original EMNAS baseline~\cite{adde2025emnas} and do not compare against the standalone hyperparameter tuning baseline from previous work~\cite{10.1007/978-3-031-82484-5_3}, as those models were trained for the full number of RL epochs and are incompatible with the low-fidelity setting used here. Finally, the winning architecture discovered through the dual-branch search is fully trained and compared with the results of the NAS-only and HPO-only baselines to evaluate the overall performance gains achieved by joint optimization.

\section{Experimental Results}
Table~\ref{tab:exp_summary} summarizes the performance of the proposed HPNAS framework compared to our previous EMNAS approach across six experiments with varying generation and population settings. The results are reported in terms of final PPO cumulative reward, total parameter count (in millions) and FLOPs (gigaFLOPs). All reported rewards correspond to the final cumulative episode return achieved by the best-performing configuration in each experiment.

\begin{table}[h]
\caption{Comparison of EMNAS and HPNAS across multiple experiments}
\label{tab:exp_summary}
\begin{tabular*}{\textwidth}{@{\extracolsep\fill}llcccccc}
\toprule
& & \multicolumn{2}{c}{Reward} & \multicolumn{2}{c}{Params (M)} & \multicolumn{2}{c}{FLOPs (G)} \\
\cmidrule(lr){3-4}\cmidrule(lr){5-6}\cmidrule(lr){7-8}
Exp & Gen / Pop & NAS & HPNAS & NAS & HPNAS & NAS & HPNAS \\
\midrule
1 & 6 / 6   & \textbf{538} & 233  & 1.29 & \textbf{1.27} & 1.26 & \textbf{1.25} \\
2 & 10 / 4  & \textbf{585} & 415  & 1.34 & \textbf{0.94} & 1.32 & \textbf{0.93} \\
3 & 10 / 10 & 569 & \textbf{621}  & 0.94 & \textbf{0.87} & 0.97 & \textbf{0.84} \\
4 & 10 / 15 & \textbf{753} & 745  & 1.45 & \textbf{0.77} & 1.44 & \textbf{0.76} \\
5 & 15 / 15 & 584 & \textbf{682}  & \textbf{0.60} & 0.84 & \textbf{0.52} & 0.82 \\
6 & 15 / 20 & 604 & \textbf{787}  & \textbf{0.51} & 0.53 & \textbf{0.42} & 0.44 \\
\botrule
\end{tabular*}
\end{table}

\noindent We observe that in smaller scale experiments (e.g., Exp~1 and 2), results are more scattered for the combined approach. This is likely due to the limited number of generations despite the larger HPNAS search space, which makes convergence more difficult. In some cases, NAS achieves higher reward, likely due to early discovery of promising architectures. However, these often lead to larger models, suggesting overparameterization when hyperparameters are not co-optimized. For instance, in Exp. 4, NAS achieves a reward of 753 but with 1.45M parameters and 1.44 GFLOPs, whereas HPNAS achieves a comparable reward of 745 with only 0.77M parameters and 0.76 GFLOPs, indicating that HPNAS finds more compact alternatives with a minimal trade-off in reward.

As the search budget increases (Exp~5 and 6), HPNAS consistently outperforms NAS in reward. While the resulting models are not always smaller, they remain comparable in size, achieving similar or slightly higher parameter counts and FLOPs. These results suggest that the HPNAS framework becomes more effective with larger populations and longer searches, better exploiting the interaction between architecture and training dynamics. This demonstrates the value of integrating HPO directly into the NAS loop, particularly in high-dimensional RL tasks.

\begin{figure}[h]
\centering
\begin{subfigure}[t]{0.49\textwidth}
    \centering
    \includegraphics[width=\linewidth]{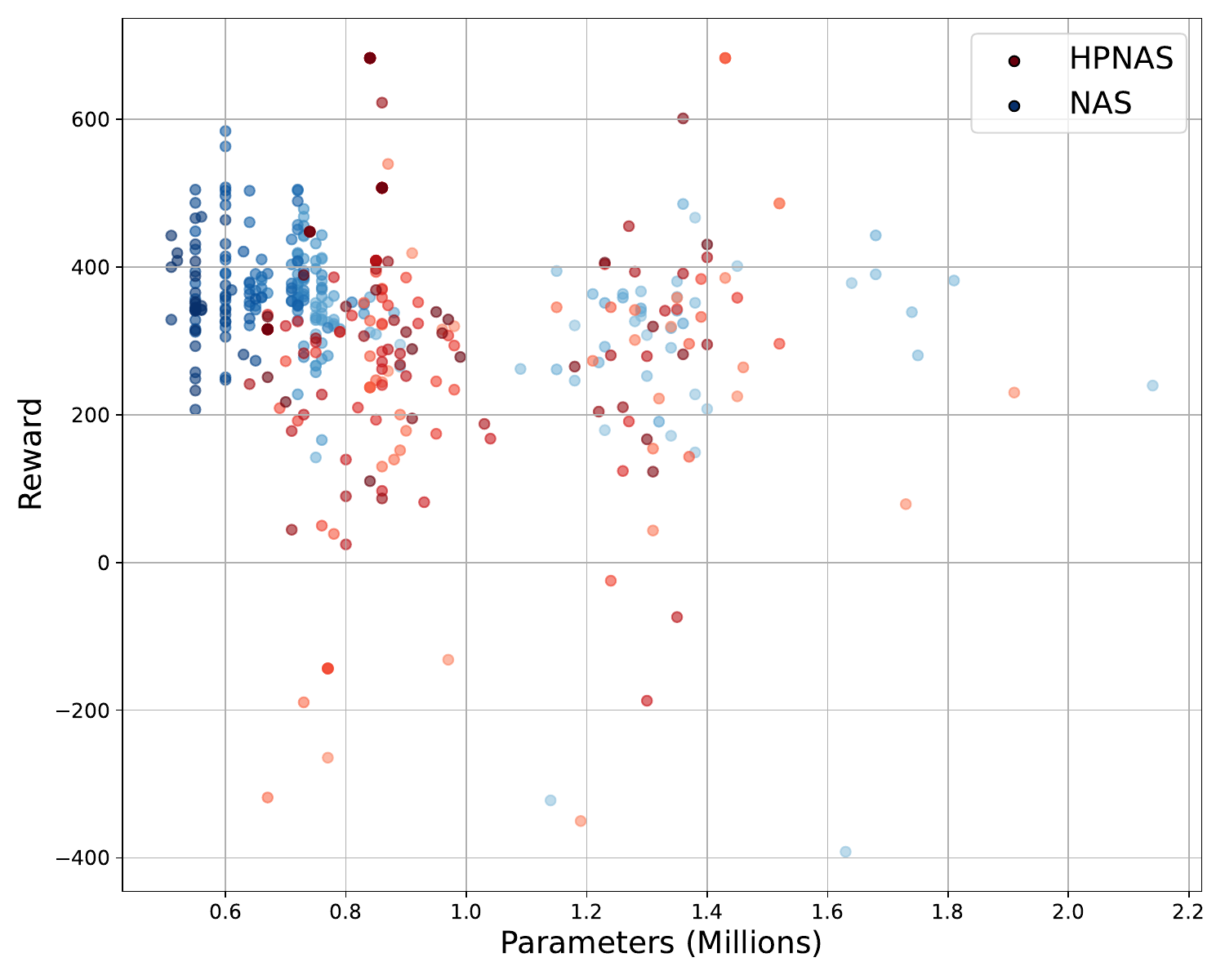}
    \caption{NAS vs HPNAS (15 Gen, 15 Pop)}
    \label{fig:evo_15x15}
\end{subfigure}
\hfill
\begin{subfigure}[t]{0.49\textwidth}
    \centering
    \includegraphics[width=\linewidth]{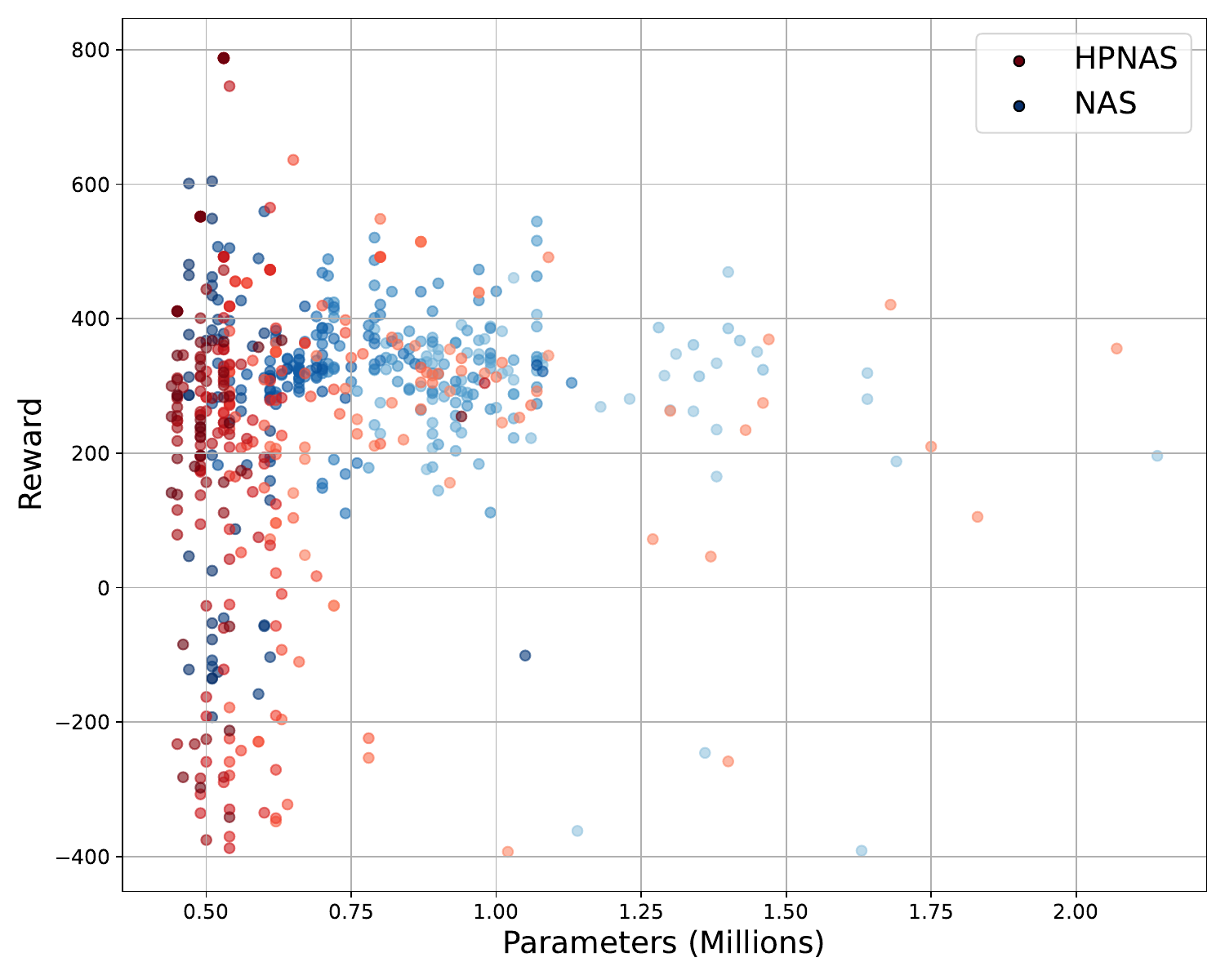}
    \caption{NAS vs HPNAS (15 Gen, 20 Pop)}
    \label{fig:evo_15x20}
\end{subfigure}
\caption{Reward–Parameter evolution across generations. Blue and red points represent NAS and HPNAS respectively, shaded from light (early generations) to dark (later generations).}
\label{fig:evo_comparison}
\end{figure}

Figure~\ref{fig:evo_comparison} shows how reward and parameter count evolve over generations in two larger experiments (15 generations with populations of 15 and 20). Since parameter count and FLOPs exhibit strong correlation across all runs, we omit FLOPs for clarity and use a 2D plot to better highlight the trade-off between performance and model efficiency. Each point represents a model, with color indicating generation (light to dark) and method (blue for NAS, red for HPNAS). Over time, HPNAS increasingly occupies the upper-left region, indicating higher rewards and smaller models. These trends highlight HPNAS's stronger exploration in early stages and its ability to converge to more efficient solutions as the search progresses.

\begin{figure}[h]
    \centering
    \includegraphics[width=0.99\linewidth]{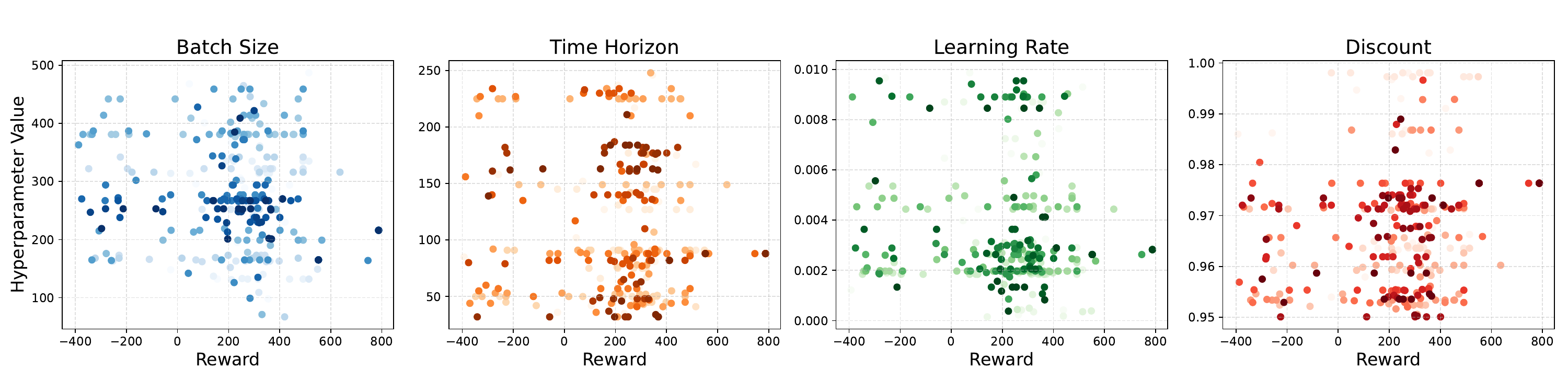} 
    \caption{Hyperparameter evolution over generations for Experiment~6 (15 generations, 20 population). Each subplot visualizes one hyperparameter, shaded from light (early generations) to dark (later generations).}
    \label{fig:hp_evolution}
\end{figure}

\noindent Figure~\ref{fig:hp_evolution} shows the evolution of four key hyperparameters across 15 generations in Experiment~6 with a population size of 20. Each subplot plots sampled hyperparameter values: batch size, time horizon, learning rate, and discount factor against their corresponding PPO rewards. Color shading reflects the generation in which the sample was evaluated, from light (early) to dark (later). The distribution shows how the search increasingly concentrates around high-performing regions. The optimal configuration converged to: batch size of 216, time horizon of 88, learning rate of $2.8 \times 10^{-3}$, and discount factor of 0.976, indicating the search effectively refined hyperparameters toward performance-optimal zones.

Finally, the RL-based autonomous driving agent was fully trained using the best-discovered architecture and its corresponding optimal hyperparameters. The resulting policy achieved a peak cumulative reward of 1205, surpassing the NAS-only configuration (1190)~\cite{adde2025emnas}, the HPO-only configuration (1190)~\cite{10.1007/978-3-031-82484-5_3}, and the manually tuned baseline (1140). This final result confirms that jointly optimizing architecture and hyperparameters yields measurable performance gains in RL-based autonomous driving tasks.

\section{Conclusion}
We proposed HPNAS, a dual-branch evolutionary framework for jointly optimizing neural architectures and hyperparameters in reinforcement learning. Experiments across multiple search budgets show that HPNAS consistently outperforms standalone NAS in both cumulative reward and model efficiency. While NAS can occasionally discover high-reward models early, it tends to overparameterize due to the absence of hyperparameter adaptation. These findings highlight the effectiveness of joint optimization in achieving compact, high-performing RL policies, particularly in larger-scale experiments.

\subsection*{Outlook}
As ongoing work, we aim to integrate a Bayesian Neural Network (BNN) surrogate model to improve sample efficiency during the final stages of evolution. The surrogate learns to predict PPO reward from joint encodings of neural architecture and hyperparameters. Once trained on NAS–HPO–reward triplets, the surrogate model enables conditional optimization~\cite{klein2017learningcurve} by inferring the optimal hyperparameter configuration that maximizes predicted reward for a given fixed architecture encoding. While the current implementation only gathers training data, future versions will activate the surrogate in the last 20\% of the search to guide sampling towards promising configurations once sufficient predictive accuracy is reached.

\bmhead{Acknowledgements}
H.G. would like to acknowledge financial support through the German Federal Ministry of Economic Affairs and Energy  through the nxtAIM consortium under grant no. 19A23014Q.

\bmhead{Disclosure of Interests}
The results, opinions and conclusions expressed in this
publication are not necessarily those of Volkswagen Aktiengesellschaft.

\bibliography{sn-bibliography}

\end{document}